\documentclass[sigconf,nonacm]{acmart}
\usepackage{url}
\usepackage{amsthm}
\usepackage{afterpage}

\usepackage[normalem]{ulem}
\useunder{\uline}{\ul}{}
\usepackage{graphicx}
\usepackage{textcomp}
\usepackage{xcolor}
\usepackage{balance} 
\usepackage{bbm}
\usepackage{amsmath,amsfonts}
\usepackage{natbib}

\usepackage{epstopdf}
\usepackage{array}
\usepackage{booktabs}

\usepackage{color}
\usepackage{xcolor}
\usepackage{colortbl}
\usepackage{xspace}
\usepackage{multirow}
\usepackage{pifont}
\usepackage{enumitem}
\usepackage{bbding}
\usepackage{hyperref}
\usepackage{makecell}
\usepackage{microtype}
\usepackage{fontawesome}
\usepackage{caption,setspace}
\usepackage{subfigure}
\usepackage[utf8]{inputenc}
\usepackage{utfsym}
\usepackage{diagbox}
\usepackage{makecell}
\AtBeginDocument{%
 }

\usepackage[linesnumbered,ruled,vlined]{algorithm2e}

\newcommand{\ssymbol}[1]{^{\@fnsymbol{#1}}}

\SetKwInput{KwInput}{Input}                
\SetKwInput{KwOutput}{Output}              
\SetKwRepeat{Do}{do}{while}

\definecolor{darkred}{rgb}{0.55, 0.0, 0.0}
\definecolor{lightred}{rgb}{0.94, 0.5, 0.5}
\definecolor{rosybrown}{rgb}{0.74, 0.56, 0.56}
\definecolor{darkgreen}{rgb}{0.0, 0.39, 0.0}
\definecolor{skyblue}{rgb}{0.56, 0.93, 0.56}
\definecolor{darkblue}{rgb}{0.0, 0.0, 0.55}
\definecolor{lightblue}{rgb}{0.68, 0.85, 0.9}
\definecolor{skyblue}{rgb}{0.53, 0.81, 0.92}

\definecolor{gray}{rgb}{0.5, 0.5, 0.5}
\definecolor{forestgreen}{rgb}{0.13, 0.55, 0.13}
\definecolor{slateblue}{rgb}{0.42, 0.35, 0.80}
\definecolor{darkgray}{rgb}{0.33, 0.33, 0.33}
\definecolor{royalblue}{rgb}{0.25, 0.41, 0.88}

\SetCommentSty{mycommfont}

\usepackage{ulem}

\keywords{Federated Graph Foundation Model, Graph Neural Networks}

\author{Zhengyu	Wu}
\affiliation{%
\institution{Beijing Institute of Technology}
\city{Beijing}
\country{China}}
\email{Jeremywzy96@outlook.com}

\author{Yinlin Zhu}
\affiliation{%
\institution{Sun Yat-sen University}
\city{Guangzhou}
\country{China}}
\email{zhuylin27@mail2.sysu.edu.cn}

\author{Xunkai Li}
\affiliation{%
\institution{Beijing Institute of Technology}
\city{Beijing}
\country{China}}
\email{cs.xunkai.li@gmail.com}

\author{Ziang Qiu}
\affiliation{%
\institution{Shandong University}
\city{Weihai}
\country{China}}
\email{ziangqiu@mail.sdu.edu.cn}

\author{Rong-Hua Li}
\authornote{Corresponding author}
\affiliation{%
\institution{Beijing Institute of Technology}
\city{Beijing}
\country{China}}
\email{lironghuabit@126.com}

\author{Guoren Wang}
\affiliation{%
\institution{Beijing Institute of Technology}
\city{Beijing}
\country{China}}
\email{wanggrbit@gmail.com}

\author{Chenghu Zhou}
\affiliation{%
\institution{Chinese Academy of Sciences}
\city{Beijing}
\country{China}}
\email{zhouch@lreis.ac.cn}

\begin{document}

\title{FedBook: A Unified Federated Graph Foundation Codebook with Intra-domain and Inter-domain Knowledge Modeling}

\renewcommand{\shortauthors}{xxx}

\begin{abstract}

    Foundation models have shown remarkable cross-domain generalization in language and vision, inspiring the development of graph foundation models (GFMs). However, existing GFMs typically assume centralized access to multi-domain graphs, which is often infeasible due to privacy and institutional constraints. Federated Graph Foundation Models (FedGFMs) address this limitation, but their effectiveness fundamentally hinges on constructing a robust global codebook that achieves intra-domain coherence by consolidating mutually reinforcing semantics within each domain, while also maintaining inter-domain diversity by retaining heterogeneous knowledge across domains. To this end, we propose FedBook, a unified federated graph foundation codebook that systematically aggregates clients’ local codebooks during server-side federated pre-training. FedBook follows a two-phase process: (1) \underline{\textit{Intra-domain Collaboration}}, where low-frequency tokens are refined by referencing more semantically reliable high-frequency tokens across clients to enhance domain-specific coherence; and (2) \underline{\textit{Inter-domain Integration}}, where client contributions are weighted by the semantic distinctiveness of their codebooks during the aggregation of the global GFM, thereby preserving cross-domain diversity. Extensive experiments on 8 benchmarks across multiple domains and tasks demonstrate that FedBook consistently outperforms 21 baselines, including isolated supervised learning, FL/FGL, federated adaptations of centralized GFMs, and FedGFM techniques.

\end{abstract}

\maketitle

\section{Introduction}
\label{sec: introduction}
    
    With the rapid progress in computational and storage technologies, foundation models have demonstrated remarkable cross-domain generalization, as exemplified by GPT-3~\cite{brown2020gpt3} in language and ViT~\cite{dosovitskiy2020vit} in vision. This success has spurred the graph learning community, leading to the emergence of various graph foundation models (GFMs)~\cite{gfm_ofa,gfm_ragraph,gfm_unigraph}. Despite their promise, existing GFMs generally assume centralized access to data, requiring graphs from multiple domains to be available to a single learning system. In practice, this assumption is rarely met, as graph data from different domains is often collected by different institutions and cannot be directly shared due to privacy regulations and competitive constraints~\cite{fu2022fgl_survey_1, zhang2021fgl_survey_2, li2024openfgl}. Consequently, it is imperative to develop effective graph foundation model training strategies for decentralized graph data resources while adequately preserving data privacy.

\begin{figure}[t]
      \includegraphics[width=0.49\textwidth]{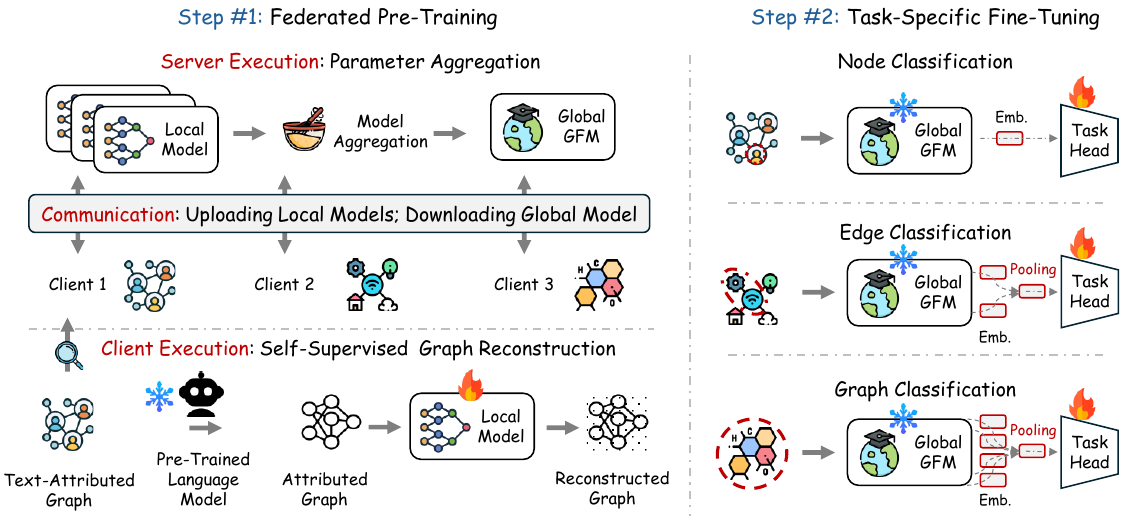}
      \captionsetup{skip=2.5pt, font={small}}
      \caption{Illustration of the FedGFM paradigm, including Federated Pre-Training and Task-Specific Fine-Tuning phases.}
      \label{fig: paradigm}
      \vspace{-0.1cm}
    \end{figure}

    To this end, recent study~\cite{zhu2025fedgfm} proposes the \textit{Federated Graph Foundation Model} (FedGFM), which combines the privacy-preserving advantages of federated graph learning (FGL) with the cross-domain generality of GFMs, enabling unified graph representation learning across domains and tasks under realistic data-silo scenarios. As illustrated in Fig.~\ref{fig: paradigm}, the FedGFM paradigm follows a two-step pipeline, including (1) \underline{\textit{Federated Pre-Training.}} Clients update local GFMs on private graphs, while a server aggregates these updates to obtain a global model that captures transferrable patterns without exposing sensitive data; and (2) \underline{\textit{Task-specific Fine-tuning.}} The learned global model serves as a GFM and is combined with task-specific heads for further fine-tuning on downstream graph tasks. Notably, graph vector quantization-masked auto-encoder (gVQ-MAE) has emerged as a suitable backbone for FedGFM, as it effectively encodes graph structures and text attributes into discrete representations~\cite{gfm_gft}, while its lightweight architecture naturally alleviates communication overhead during federated pre-training.

    Building on these insights, we argue that establishing an effective FedGFM technique essentially hinges on learning a global gVQ-MAE, whose codebook should satisfy the following \textbf{key properties}: (1) \underline{\textit{Intra-domain Coherence.}} The global codebook should integrate knowledge from multiple clients within the same domain in a way that domain-specific semantics become mutually reinforcing; and (2) \underline{\textit{Inter-domain Diversity.}} The global codebook should preserve heterogeneous semantics across domains, ensuring that knowledge from different domains is jointly represented rather than overlooked. Notably, as the first attempt within the FedGFM paradigm, FedGFM+~\cite{zhu2025fedgfm} introduces a carefully designed initialization scheme to provide the global codebook with a favorable starting point prior to federated pre-training. While this design enhances the initial representational quality, the subsequent pre-training phase still adopts a simplistic parameter aggregation strategy. As a result, the broader challenge of systematically designing effective global codebook aggregation strategies remains unexplored.

    \vspace{0.05cm}
    
    To this end, we introduce \textbf{FedBook}, a unified \textbf{\underline{fed}}erated graph foundation code\textbf{\underline{book}} within the FedGFM paradigm, which consolidates multi-domain knowledge from clients’ local codebooks during server-side federated pre-training. FedBook proceeds in two phases aligned with the design goals outlined earlier: 
    (1) \textbf{Intra-domain Collaboration Phase.} To ensure \underline{\textit{Intra-domain Coherence}}, FedBook aggregates knowledge from clients within the same domain. Each token in a local codebook is treated as a discrete semantic unit, with its frequency indicating the reliability of its semantics. During aggregation, low-frequency tokens are refined by referencing semantically similar high-frequency tokens from other clients, thereby improving coherence across clients. 
    (2) \textbf{Inter-domain Integration Phase.} To preserve \underline{\textit{Inter-domain Diversity}}, FedBook assesses the semantic distinctiveness of each client’s codebook and weights their contributions accordingly. Clients with more distinctive semantics exert greater influence on the global codebook, ensuring their knowledge is preserved rather than diluted.

    \vspace{0.05cm}
    
    \textbf{Our Contributions.} (1) \textit{\underline{New Perspective.}} We are the first to study FedGFM from the lens of parameter aggregation, highlighting how the global codebook can effectively integrate multi-domain semantics. (2) \textit{\underline{Novel Method.}} We introduce FedBook, a two-phase aggregation FedGFM framework with coherent and diverse semantic representation. (3) \textit{\underline{State-of-the-art Performance.}} We conduct experiments on 8 diverse benchmarks across multiple domains and tasks, where FedBook consistently surpasses 21 baselines from isolated supervised learning, FL/FGL, federated adaptations of centralized GFMs, and FedGFM.

\section{Preliminaries}
\label{sec: preliminaries}

    \noindent \textbf{Text-Attributed Graphs (TAGs).} 
    We consider a text-attributed graph $G=(\mathcal{V},\mathcal{E})$, where $\mathcal{V}$ and $\mathcal{E}$ denote the node and edge sets, respectively. 
    Each node $v_i \in \mathcal{V}$ or edge $e_i \in \mathcal{E}$ may be associated with textual descriptions, which are encoded into semantic vector representations using pre-trained language models (e.g., Sentence-BERT~\cite{reimers2019sentence_bert}). 
    Supervision signals can be specified at different granularities depending on the downstream task: node level (e.g., node classification), edge level (e.g., link prediction), or graph level (e.g., graph classification). Notably, within the FedGFM paradigm, each client retains its own pre-trained language model instance with an identical version for feature extraction of TAGs, ensuring feature spaces remain aligned across heterogeneous domains and tasks.

    \vspace{0.1cm}
    \noindent \textbf{Graph Vector Quantization-Masked Auto-Encoder (gVQ-MAE).} 
    Various GFMs employ gVQ-MAEs as their backbone, which jointly encode structural topology and textual attributes into a discrete embedding space with well-defined semantic boundaries, making them particularly well-suited for multi-domain pre-training. Specifically,
    (1) \textit{$\mathcal{G}^\prime=(\mathcal{V},\mathcal{E},\mathcal{X}) \to$ \textbf{Encoder} $\to$ Embeddings}: 
    To ensure generality across arbitrary inputs, the Encoder may be instantiated as any reasonable GNN that incorporates both node- and edge-level information to produce embeddings $z\in\mathbb{R}^{d}$.
    (2) \textit{Embeddings $\to$ \textbf{Codebook} $\to$ Quan. Emb.}: 
    To impose discrete semantics, the Codebook $\mathcal{C}$ maps continuous embeddings $z$ into quantized embeddings $e\in\mathbb{R}^{d}$ (Quan. Emb. $z_q\in\mathbb{R}^{d}$) via similarity-based vector quantization:
    \begin{equation}
    \label{eq: lookup}
    z_q \leftarrow e_j,\quad j=\arg\min_{e_i\in\mathcal{C}}\Vert z-e_i\Vert_2,\quad \mathcal{C}=\{e_1,e_2,\dots,e_K\}.
    \end{equation}
    (3) \textit{Quan. Emb. $\to$ \textbf{Decoder} $\to$ $\mathcal{G}_r^\prime=(\mathcal{V},\mathcal{E}_r,\mathcal{X}_r)$}: 
    For self-supervised training, gVQ-MAEs adopt an autoencoder framework in which gradients are derived from the discrepancy between the reconstructed graph $\mathcal{G}_r^\prime$ and the original input $\mathcal{G}^\prime$, thereby updating both the Encoder and Codebook. 
    Notably, the trainable components consist of the Encoder’s weight matrices and the discrete embeddings $\{e_1,\dots,e_K\}$ in the Codebook, together forming the learnable GFM embedding function parameterized by $f_\theta$. 
    To enable end-to-end optimization, the straight-through estimator (STE)~\cite{bengio2013estimating_vqvae_ste} is employed to approximate gradients by bypassing the non-differentiable quantization step. Formally, the overall pre-training loss is defined as:
    \begin{equation}
    \label{eq: pretrain loss}
    \mathcal{L}_{\text{pretrain}} = \mathcal{L}_{\text{feat}} + \mathcal{L}_{\text{topo}} 
    + \frac{1}{n}\sum_{i=1}^{n}\|\text{sg}[z_i] - z_{q_i}\|_2^2
    + \frac{1}{n}\sum_{i=1}^{n}\|z_i - \text{sg}[z_{q_i}]\|_2^2 ,
    \end{equation}
    where $\text{sg}[\cdot]$ denotes the stop-gradient operator, $z_i$ is the embedding of node $i$, and $z_{q_i}$ is its quantized embedding retrieved from the codebook. Moreover, the feature reconstruction and topology reconstruction terms are given by:
    \begin{equation}
    \mathcal{L}_{\text{feat}} = \frac{1}{n}\sum_{i=1}^{n} \Big(1 - \frac{x_i^\top \hat{x}_i}{\|x_i\| \cdot \|\hat{x}_i\|}\Big)^{\gamma}, 
    \qquad
    \mathcal{L}_{\text{topo}} = \|A - \sigma(\hat{X}\hat{X}^\top)\|_2^2 ,
    \end{equation}
    where $x_i$ and $\hat{x}_i$ are the original and reconstructed node attributes, respectively; $\gamma$ is a scaling factor; $A$ is the adjacency matrix ($A_{ij}=1$ if $e_{ij}\in\mathcal{E}$ otherwise $0$); and $\sigma(\cdot)$ denotes the sigmoid function.

    \vspace{0.15cm}
    \noindent \textbf{Problem Formalization of FedGFM.} Consider a federated setting with a trusted central server and $K$ clients. Each client $i \in \{1,\dots,K\}$ owns a private graph collection $\mathcal{S}_i$, where $|\mathcal{S}_i| = 1$ for subgraph-level decentralization. The goal of FedGFM is to collaboratively train a graph foundation model while preserving data locality.  The paradigm follows a \textit{federated pre-training, and task-specific fine-tuning} process.  Specifically, for the \textbf{Federated Pre-Training} phase, at each communication round $r=1,\dots,R$, the following procedure is executed: (1) Initialization: At round $r=1$, the server initializes the global model with parameters $\mathbf{W}^g$ and sends them to all clients, setting $\mathbf{W}^i \leftarrow \mathbf{W}_g, \;\forall i$.  (2) Local Updates: Each client $i$ trains its local model on private graph data $G_i \in \mathcal{S}_i$ by minimizing a self-supervised loss $\mathcal{L}(G_i;\mathbf{W}_i)$, and updates the model parameters via stochastic gradient descent: $\mathbf{W}^i \leftarrow \mathbf{W}^i - \eta \nabla \mathcal{L}(G_i;\mathbf{W}^i)$.  (3) Global Aggregation: After local updates, each participating client sends its updated parameters $\mathbf{W}^i$ to the server. The server aggregates them proportionally to the number of local training instances $M_k$, formulated as $
    \mathbf{W}^g \leftarrow \sum_{i=1}^K \frac{M_i}{M}\,\mathbf{W}^i \;
    \text{with} \;  M = \sum_{i=1}^K M_i$, and then broadcasts the updated global parameters $\mathbf{W}^g$ back to the clients. This iterative procedure repeats until the final round $R$, resulting in a pre-trained global graph foundation model. For \textbf{Task-specific Fine-tuning} phase, the server distributes the final global model $\mathbf{W}^g$ to all clients. Each client loads the shared backbone as a frozen encoder and fine-tunes lightweight, task-specific heads using the available supervision signals. Specifically, in this paper, the task-specific heads consist of a prototype classifier and a linear classifier, consistent with previous studies~\cite{gfm_gft, zhu2025fedgfm}. We optimize the cross-entropy loss between the predictions and the ground truth to update both classifiers, formulated as follows:
    \begin{equation} 
        \begin{aligned} 
        p^{\text{proto}}(y=i|\mathbf{z}) &= \frac{\exp(-\text{sim}(\mathbf{z}, \mathbf{p}_i))}{\sum_c\exp(-\text{sim}(\mathbf{z}, \mathbf{p}_c))}, \\ 
        p^{\text{linear}}(y=i|\mathbf{z}) &= \frac{\exp(\text{lin}^i(\mathbf{q}))}{\sum_c\exp(\text{lin}^c(\mathbf{q}))}, 
        \end{aligned}
    \end{equation}
    \noindent where $\mathbf{z}$ denotes the quantized embedding of the test sample (node, edge, or graph), $\mathbf{p}_i$ denotes the prototype of class $i$ derived from the training set, and $\text{lin}^i(\mathbf{q})$ denotes the $i$-th soft logit. During inference, the predictions from the prototype and linear classifiers are combined to produce the final output.

\begin{figure*}[t]
  \includegraphics[width=0.998\textwidth]{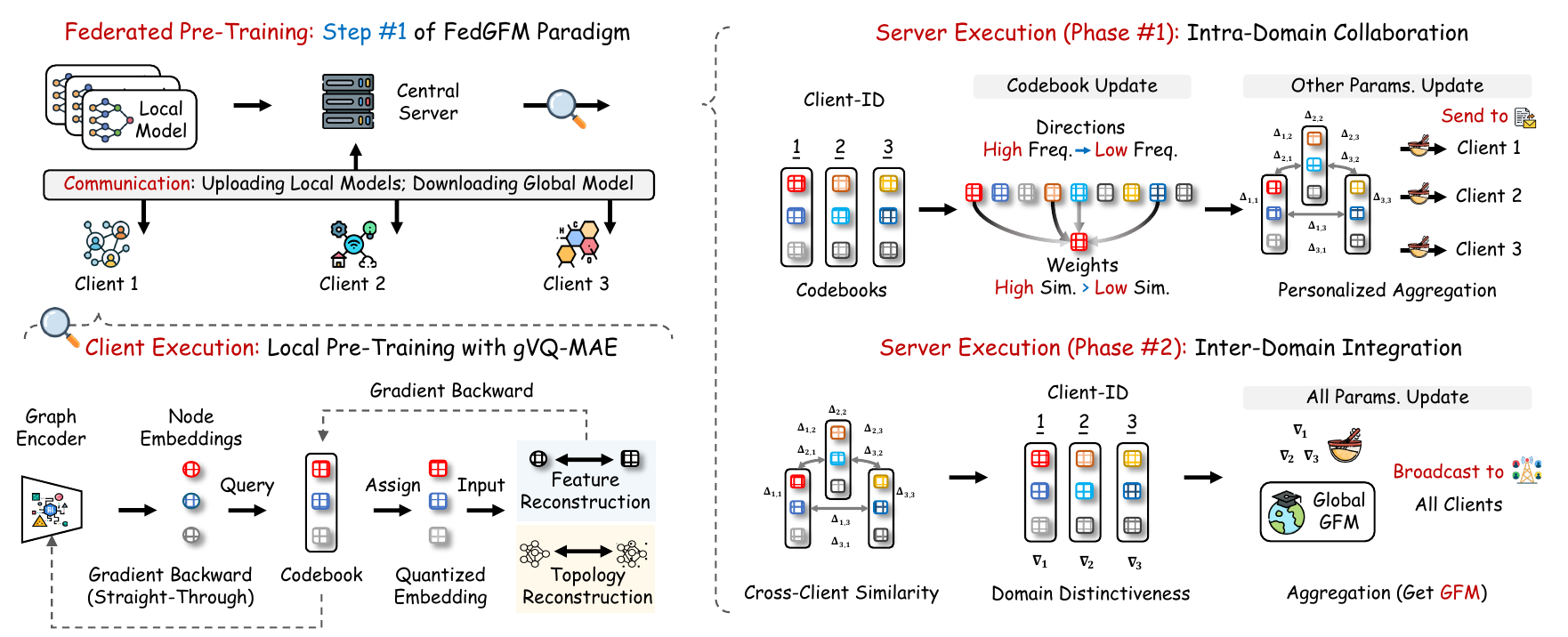}
  \captionsetup{skip=2.5pt, font={small,stretch=1}}
  \caption{Overview of the proposed FedBook framework under the FedGFM paradigm. In each communication round, clients perform local pre-training with their gVQ-MAE on local graph data, while the server sequentially conducts intra-domain collaboration followed by inter-domain integration to preserve intra-domain coherence and inter-domain diversity, ultimately yielding an effective global GFM.}
\label{fig: fedbook}
\vspace{0.15cm}
\end{figure*}

\section{Methodology}
\label{sec: methodology}

    In this section, we introduce FedBook, the unified federated graph foundation codebook under the FedGFM paradigm, as depicted in Fig.~\ref{fig: fedbook}. FedBook is designed to enhance server-side aggregation by consolidating multi-domain knowledge from local codebooks into a global one. It specifically functions through two complementary phases: one is \underline{\textit{Intra-domain Collaboration}}, which improves coherence among clients within the same domain, and the other is \underline{\textit{Inter-domain Integration}}, which maintains semantic diversity across different domains. Through these two mechanisms, FedBook effectively coordinates intra-domain and inter-domain knowledge modeling to learn a powerful global GFM.

\subsection{Phase 1: Intra-domain Collaboration}
\label{sec: Intra-Domain}

    As discussed in Section~\ref{sec: introduction}, the effectiveness of FedGFM fundamentally relies on constructing a powerful global gVQ-MAE. A key requirement for its codebook is \underline{\textit{Intra-domain Coherence}}, i.e., the ability to utilize knowledge from the same domain such that semantics are mutually reinforced. This phase is thus designed by answering three questions: \textbf{Q1}: How do we identify semantically similar knowledge across clients? \textbf{Q2}: How to determine the more reliable representation to serve as the anchor for intra-domain semantic alignment? and \textbf{Q3}: How to update parameters other than the codebook of gVQ-MAE?

\vspace{0.1cm}

    \noindent \textbf{Token as the Semantic Unit.} To answer \textbf{Q1}, we draw on the design philosophy of gVQ-MAE, which encodes domain knowledge as discrete representations, where each token is treated as an independent semantic unit. Based on this insight, we propose that similar tokens correspond to similar knowledge between clients in the FedGFM paradigm. Specifically, consider the local codebooks of client $a$ and client $b$ with tokens $\{\phi^a_i\}_{i=1}^{N}$ and $\{\phi^b_j\}_{j=1}^{N}$ ($N$ is the number of tokens), we denotes the similarity between client $a$'s $i$-th token and client $b$'s $j$-th token as $S_{i,j}^{a,b}$, formulated as:
    \begin{equation}
    \label{eq: token_similarity}
    \mathbf{S}_{i,j}^{a,b}=\frac{\phi_i^a\phi_j^b}{|\!|\phi_i^a|\!|_2|\!|\phi_j^b|\!|_2}.
    \end{equation}

\vspace{0.1cm}
    \noindent \textbf{Frequency-guided Semantic Alignment}. To answer \textbf{Q2}, even clients from similar fields may learn category semantics to different degrees due to data heterogeneity, that is, the local sample distribution of each client varies significantly. Under the FedGFM paradigm, we aim to transfer knowledge from high-quality to low-quality clients by modeling knowledge quality as the token access frequency during training, where higher frequency indicates more samples mapped to the token. Specifically, let $\delta_i^a$ denote the $i$-th token of client $a$ and $\delta_j^b$ the $j$-th token of client $b$, the alignment weight from $\delta_j^b$ to $\delta_i^a$ is defined as:
    \begin{equation}
    \label{eq: masked_similarity}
    \mathbf{\tilde{S}}_{i\leftarrow j}^{a\leftarrow b} =
    \begin{cases}
    \mathbf{S}_{i,j}^{a,b}, & \text{if } \delta_j^b \geqslant \delta_i^a, \\
    -\infty, & \text{otherwise}.
    \end{cases}
    \end{equation}
    Finally, for each token $\phi_i^a$, we perform semantic alignment with all higher-frequency tokens across all clients. The aligned token representation is computed as a weighted aggregation of candidate tokens, where the weights are obtained by normalizing the masked similarities. Specifically, we use $\hat{\phi}_i^a$ to represent the updated representation of token $\phi_i^a$, which is computed as follows:
    \begin{equation}
    \label{eq: codebook_update_phase_1}
    \hat{\phi}_i^a
    = \lambda \phi^a_i + (1-\lambda)
    \sum_{b=1}^{K}\sum_{j=1}^{N} 
    \frac{\exp\big(\mathbf{\tilde S}_{i\leftarrow j}^{\,a\leftarrow b}\big)}
    {\sum_{b'=1}^{K}\sum_{j'=1}^{N} \exp\big(\mathbf{\tilde S}_{i\leftarrow j'}^{\,a\leftarrow b'}\big)}
    \, \phi_j^b,
    \end{equation}
    \noindent where $\lambda$ denotes the trade-off parameter, $K$ denotes the total number of clients. By setting the masked similarity $\mathbf{\tilde S}_{i\leftarrow j}^{,a\leftarrow b}$ to negative infinity for lower-frequency tokens, the aggregation is limited to contributions from higher-frequency tokens.

\vspace{0.1cm}
    \noindent \textbf{Personalized Model Aggregation.} To answer \textbf{Q3}, we calculate the semantic similarity between clients and customize the aggregation process for each client to meet the intra-domain collaboration goal of this phase. Specifically, consider the local codebooks of client $a$ and client $b$ with tokens $\{\phi^a_i\}_{i=1}^{N}$ and $\{\phi^b_i\}_{i=1}^{N}$, we model the semantic similarity between two clients as the average maximum token-wise similarity of their codebooks, denoted as $\Delta^{a,b}$, formulated as:
    \begin{equation}
    \label{eq: client_similarity}
    \Delta^{a,b} 
    = \frac{1}{N} \sum_{i=1}^{N} \max_{1 \leqslant j \leqslant K} \mathbf{S}_{i,j}^{a,b}.
    \end{equation}
    Notably, the use of the maximum similarity value $\max_{1\leqslant j \leqslant K}\mathbf{S}_{i,j}^{a,b}$ is motivated by its ability to identify the most semantically relevant token in client $b$'s codebook for each token in client $a$'s codebook, thus capturing the strongest semantic correspondence between clients and support the subsequent personalized aggregation.

    Afterward, the parameters other than the codebook of gVQ-MAE for client $a$ can be updated as follows:
    \begin{equation}
    \label{eq: other_param_update_phase_1}
    \mathbf{W}^a = \sum_{b=1}^K \frac{\exp(\Delta^{a,b})}{\sum_{b'=1}^N \exp(\Delta^{a,b'})} \mathbf{W}^b,
    \end{equation}
    \noindent where $\mathbf{W}^i$ denotes the parameters other than the codebook of gVQ-MAE (e.g., encoder and decoder) for the $i$-th client.

    Finally, the server will send personalized gVQ-MAE parameters for each client, which then updates its local parameters accordingly before proceeding to the next round of local pre-training.

\subsection{Phase 2: Inter-domain Integration}
\label{sec: Inter-Domain}

As discussed in Section~\ref{sec: introduction}, another key property of the gVQ-MAE codebook in the FedGFM paradigm is \underline{\textit{Inter-domain Diversity}}, i.e., the global codebook should preserve heterogeneous semantics across domains, ensuring that domain-specific knowledge is jointly represented rather than overlooked. Guided by this principle, this phase is designed to address two questions: \textbf{Q1}: How to characterize the distinctiveness of each client’s local graph domain? and \textbf{Q2}: How to aggregate local codebooks for multiple clients into the global GFM while preserving unique domain knowledge?

\vspace{0.1cm}    
\noindent \textbf{Domain Distinctiveness.} To address \textbf{Q1}, we posit that clients operating in distinct domains will develop local codebooks with unique semantic features. Building on the client-wise similarity defined in Eq.~\ref{eq: other_param_update_phase_1}, we quantify the domain distinctiveness of client $a$ as $\nabla^a$, which can be calculated as follows:
\begin{equation}
\label{eq: domain distinctiveness}
\nabla^a = 1 - \frac{1}{K}\sum_{b=1}^K \Delta^{a,b}.
\end{equation}

\noindent \textbf{Global gVQ-MAE Aggregation}. To answer \textbf{Q2}, our motivation is to assign higher aggregation weights to clients with high domain uniqueness, as their knowledge cannot be supplemented by clients similar to other domains. Specifically, we use $\mathbf{\hat{W}^{g}}$ to denote the global GFM, which is computed as follows:
\begin{equation}
\label{eq: global GFM aggregation}
\mathbf{\hat{W}^g} = \sum_{a=1}^K \frac{\exp(\nabla^a)}{\sum_{a'=1}^K \exp(\nabla^{a'})} \mathbf{\hat{W}}^a,
\end{equation}
\noindent where $\mathbf{\hat{W}}^i$ denotes all parameters of local gVQ-MAE (i.e., $\mathbf{\hat{W}}^i=\mathbf{W}^i \cup \{\phi^i_j\}_{j=1}^N$) for the $i$-th client. The aggregated global gVQ-MAE is then broadcast to all clients for the next round of local pre-training.

In summary, after multiple rounds of \underline{\textit{Phase~1}}, each client's gVQ-MAE parameters are iteratively refined by assimilating knowledge from other similar clients, yielding more robust domain-specific semantics. Subsequently, the iterative process of \underline{\textit{Phase~2}} further effectively integrates the heterogeneous knowledge, resulting in a generalized global GFM. The complete implementation procedure of the proposed FedBook framework is outlined in Algorithm~\ref{alg: fedbook}.

\begin{algorithm}[htbp]
\fontsize{8pt}{4pt}\selectfont
\SetAlgoLined
\LinesNumbered
\caption{Clients and Server Execution of FedBook}
\label{alg: fedbook}
\SetKwFunction{FServerPO}{S\_Exec\_P1}
\SetKwFunction{FServerPT}{S\_Exec\_P2}
\SetKwFunction{FClient}{C\_Exec}
\SetKwProg{Fn}{Function}{:}{end}
\small

\KwInput{Communication rounds $R_1$, $R_2$ for Phase~1 and Phase~2; Local pre-training epochs $E$; Number of clients $K$; Initialized global GFM with parameters $\hat{\mathbf{W}}^g = \mathbf{W}^g \cup \{\phi_i^g \}_{i=1}^N$}.
\KwOutput{Learned global GFM with parameters $\hat{\mathbf{W}}^g$}

\ForPar{each client $i=1,...,K$}{
    $\{\phi_j^{i\text{-cache}}\}_{j=1}^N \gets \{\phi_j^g\}_{j=1}^N$ \tcp*{codebook cache init.}
    $\mathbf{W}^{i\text{-cache}} \gets \mathbf{W}^g$ \tcp*{other param. cache init.}
}

\For{each communication round $r$ \textup{in} $1,...,R_1+R_2$}{
    \tcc{Client Execution}
    \ForPar{each client $i=1,...,K$}{
        $\{\phi_j^{i}\}_{j=1}^N, \mathbf{W}^{i} \gets $\FClient{$i, \{\phi_j^{i\text{-cache}}\}_{j=1}^N, \mathbf{W}^{i\text{-cache}}$}; 
    }
    \tcc{Server Execution (Phase 1)}
    \If{$1 \leqslant r \leqslant R_1$}{
        $\{\phi_j^{i\text{-cache}}\}_{j=1}^N, \mathbf{W}^{i\text{-cache}} \gets$ \FServerPO{$\{\{\phi_j^{i}\}_{j=1}^N, \mathbf{W}^{i}\}_{i=1}^K$};
    }
    
    \tcc{Server Execution (Phase 2)}
    \If{$R_1 < r \leqslant R_2$}{
        $\{\phi_j^{i\text{-cache}}\}_{j=1}^N, \mathbf{W}^{i\text{-cache}} \gets$ \FServerPT{$\{\{\phi_j^{i}\}_{j=1}^N, \mathbf{W}^{i}\}_{i=1}^K$};
    }
}

\Fn{\FClient{$i, \{\phi_j^{i\text{-cache}}\}_{j=1}^N, \mathbf{W}^{i\text{-cache}}$}}{
    {$\{\phi_j^{i}\}_{j=1}^N \gets \{\phi_j^{i\text{-cache}}\}_{j=1}^N$}\tcp*{Codebook init.}
    {$\mathbf{W}^{i} \gets \mathbf{W}^{i\text{-cache}}$}\tcp*{Other params. init.}

    \For{each local training epoch $e$ \textup{in} $1, ..., E$}{
        Local pre-training to update local gVQ-MAE \tcp*{Eq.~(\ref{eq: pretrain loss})}
    }
    \Return{$\{\phi_j^{i}\}_{j=1}^N, \mathbf{W}^{i}$}\tcp*{Upload codebook and other params.}
}

\tcc{Server Execution: Phase 1 - Intra-domain Collaboration}
    \Fn{\FServerPO{$\{\{\phi_j^{i}\}_{j=1}^N, \mathbf{W}^{i}\}_{i=1}^K$}}{
    Compute token-wise similarity $\mathbf{S}$ \tcp*{Eq.~(\ref{eq: token_similarity})}
    
    Compute alignment weight $\tilde{S}$ \tcp*{Eq.~(\ref{eq: masked_similarity})}

    Compute client-wise similarity $\Delta$ \tcp*{Eq.~(\ref{eq: client_similarity})}
    
    \ForPar{each client $i=1,...,K$}{
        Update codebook $\{\phi_j^{i}\}_{j=1}^N$ \tcp*{Eq.~(\ref{eq: codebook_update_phase_1})}
        Update other local params $\mathbf{W}^{i}$ \tcp*{Eq.~(\ref{eq: other_param_update_phase_1})}
    }
    \Return{$\{\{\phi_j^{i}\}_{j=1}^N, \mathbf{W}^{i}\}_{i=1}^K$}
}   
    
\tcc{Server Execution: Phase 2 - Inter-domain Integration}
    \Fn{\FServerPT{$\{\{\phi_j^{i}\}_{j=1}^N, \mathbf{W}^{i}\}_{i=1}^K$}}{
        Compute domain distinctiveness $\nabla$ \tcp*{Eq.~(\ref{eq: domain distinctiveness})}
        Params. aggregation for global gVQ-MAE $\hat{\mathbf{W}}^g$  \tcp*{Eq.~(\ref{eq: global GFM aggregation})} 
        \Return{$\hat{\mathbf{W}}^g$} \tcp*{Broadcast global params. to all clients}
    }   
\end{algorithm}

\begin{table*}[t]
\setlength{\abovecaptionskip}{0.1cm}
\renewcommand{\arraystretch}{1.5}
\caption{Statistical summary of the experimental datasets, where multi-label classification datasets are denoted by the symbol `*'.}
\footnotesize 
\label{tab: datasets}
\resizebox{\linewidth}{20mm}{
\setlength{\tabcolsep}{6mm}{
    \begin{tabular}{ccccccccc}
      \toprule
      \textbf{Dataset}  & \textbf{Domain} & \textbf{Task Level} & \textbf{\#Graphs} & \textbf{Avg. \#Nodes} & \textbf{Avg. \#Edges} & \textbf{\#Classes} & \textbf{train/val/test} \\ \midrule
      Cora    & Citation        & Node          & 1                  & 2,708                 & 10,556                & 7  & 5\%/20\%/40\%                 \\
      PubMed   & Citation        & Node          & 1                  & 19,717                & 44,338                & 3    & 60\%/20\%/20\%               \\
      OGB-arxiv    & Citation        & Node          & 1                  & 169,343               & 1,166,243             & 40     & 80\%/10\%/10\%             \\
      WikiCS   & Hyper link        & Node          & 1                  & 11,701                & 216,123               & 10      & 80\%/10\%/10\%            \\
      FB15K237 & Knowledge       & Link          & 1                  & 14,541                & 310,116               & 237             & 80\%/10\%/10\%    \\
      WN18RR   & Knowledge       & Link          & 1                  & 40,943                & 93,003                & 11               & 80\%/10\%/10\%   \\
      PCBA$^*$     & Molecule        & Graph         & 437,929            & 26.0                  & 28.1                  & 128   & 80\%/10\%/10\% \\
      HIV$^*$      & Molecule        & Graph         & 41,127             & 25.5                  & 27.5                  & 2  & 80\%/10\%/10\% \\
      \bottomrule
    \end{tabular}
}}
\end{table*}

\section{Experiments}

    In this section, we present a comprehensive evaluation of our proposed FedBook framework. We begin by detailing the experimental setup (Sec~\ref{sec: experimental setup}). Subsequently, our evaluation is structured around addressing the following key research questions: \textbf{Q1}: After task-specific fine-tuning, does FedBook consistently outperform state-of-the-art baselines across all tasks in the FedGFM scenario (Sec.~\ref{sec: performance comparison})?
    \textbf{Q2}: What is the individual contribution of each core component in FedBook to the overall performance (Sec.~\ref{sec: ablation study})? \textbf{Q3}: How robust is FedBook to changes in hyperparameters (Sec.~\ref{sec: sensitivity analysis})? \textbf{Q4}: Does FedBook outperform federated adaptations of centralized GFMs in few-shot learning (Sec.~\ref{sec: few shot})? \textbf{Q5}: How dose FedBook perform in terms of efficiency (Sec.~\ref{sec: efficiency})?

\subsection{Experimental Setup}
\label{sec: experimental setup}
    We first describe the datasets, data simulation strategy, baselines, and evaluation metrics, and then specify the model architectures. Additional experimental details are provided in Appendix~\ref{appendix: exp setting}.

\vspace{0.1cm}
\noindent \textbf{Datasets.}
    We evaluate FedBook on 8 benchmark graph datasets across 6 domains and 3 tasks, including: (1) Node classification datasets (Cora, PubMed~\cite{Yang16cora}, OGB-arxiv~\cite{hu2020ogb} and WikiCS~\cite{mernyei2020wikics}); (2) Edge classification datasets (FB15K237~\cite{toutanova2015fb15k237} and WN18RR~\cite{dettmers2018dataset_wn18rr}); (3) Graph classification datasets (PCBA and HIV~\cite{wu2018dataset_moleculenet}). Detailed statistical information of these datasets is presented in Table.~\ref{tab: datasets}.

\vspace{0.1cm}
\noindent \textbf{Data Simulation.} To better approximate real-world distributed graph sources, each dataset in Table~1 was further split into multiple subsets. Partitioning strategies followed task-specific conventions widely used in FedGFM~\cite{zhu2025fedgfm} and FGL studies~\cite{zhang2021fedsage, baek2022fedpub, wan2024fgl_fggp}. For node and edge classification, we applied the Metis algorithm~\cite{karypis1998metis} to detect communities and assigned them to clients. For graph classification, since each graph carries multiple labels, we simply partitioned the graphs randomly and evenly by label count. Unless otherwise stated, each dataset was distributed across 3 clients, yielding a total of 24 clients in the simulated FedGFM scenario.

\vspace{0.1cm}
\noindent \textbf{Baselines.}
We compare FedBook with four categories of baselines: (1) Isolated supervised models, trained independently on each client (Linear, GCN~\cite{kipf2016gcn}, GAT~\cite{velivckovic2017gat}, GraphSAGE~\cite{hamilton2017graphsage}, GIN~\cite{xu2018gin}); (2) FL/FGL methods, including general-purpose approaches (FedAvg~\cite{mcmahan2017fedavg}, MOON~\cite{li2021moon}) and task-specific ones (FedSage+~\cite{zhang2021fedsage}, Fed-PUB~\cite{baek2022fedpub}, FedGTA~\cite{li2024fedgta}, FedTAD~\cite{zhu2024fedtad}, FGSSL~\cite{huangfgl_fgssl}, FGGP~\cite{wan2024fgl_fggp}, GCFL+~\cite{xie2021gcfl}, FedStar~\cite{tan2023fgl_fedstar}); (3) Federated adaptations of centralized GFM (OFA~\cite{gfm_ofa}, GFT~\cite{gfm_gft}, UniGraph~\cite{gfm_unigraph}, GQT~\cite{gfm_gqt}, GraphCLIP~\cite{gfm_graphclip}); and (4) FedGFM methods, represented by FedGFM+~\cite{zhu2025fedgfm}.

\vspace{0.1cm}
\noindent \textbf{Evaluation Metrics.} For each dataset, we report the mean and variance of the evaluation metrics computed over the test sets of its five associated clients across 10 standardized runs. We employ Accuracy (ACC) for node and edge classification tasks, and the Area Under the Receiver Operating Characteristic Curve (AUC-ROC) for graph classification. The evaluation protocol differs by method: for Isolated Supervised Learning baselines, we evaluate each client's local model directly without collaboration; for FL and FGL baselines, we evaluate client models after multiple rounds of global training; for federated centralized GFM and FedGFM baselines, we attach the task-specific heads and fine-tune the model on each client's local graph prior to evaluation.

\vspace{0.14cm}
\noindent \textbf{Model Architecture.} Our experimental setup employs distinct model architectures tailored to each baseline category:
(1) For isolated supervised models, we adopt a two-layer network with 64 hidden units;
(2) For FL/FGL methods, we employ task-specific backbone architectures unless otherwise specified by the original method: GraphSAGE is used for node and edge classification, and GIN for graph classification. Otherwise, we follow the custom architectures defined in their original papers.
(3) For federated adaptations of centralized GFM methods, we retain the backbones reported in their original papers;
(4) For FedGFM method (notably FedGFM+~\cite{zhu2025fedgfm} as the sole existing baseline), we faithfully replicate its architectural setup;
(5) Finally, for our proposed FedBook framework, we employ a gVQ-MAE backbone across both client-local and server-global models. The encoder comprises a 2-layer GraphSAGE-based graph convolutional network that jointly processes node and edge features. All layers maintain a uniform hidden dimension of 768 to align with Sentence-BERT~\cite{reimers2019sentence_bert} embeddings of graph attributes. The gVQ-MAE architecture further incorporates a codebook with 4 heads, each containing 128 learnable tokens, and utilizes a shared linear projection to aggregate multi-head outputs into the final quantized representation.

\begin{table*}[h]
    \setlength{\abovecaptionskip}{0.2cm} 
    \renewcommand{\arraystretch}{1.5} 
    \caption{Performance comparison between FedBook and 21 baselines. The globally best and category-specific second-best results are highlighted in \textcolor{red}{Red} and \textcolor{blue}{Blue}, respectively. Symbols `*' denotes the federated adaptations of centralized GFMs; `N/A' represents the task inapplicability.}
    \footnotesize 
    \label{tab: compare baseline}
    \resizebox{\linewidth}{63mm}{  
    \setlength{\tabcolsep}{2.4mm}{  
    \begin{tabular}{c|c|c|c|c|c|c|c|c}
        \toprule[1pt] 
         \textbf{Method} & \textbf{Cora} & \textbf{PubMed} & \textbf{OGB-arxiv} & \textbf{WikiCS} & \textbf{FB15K237} & \textbf{WN18RR} & \textbf{HIV} & \textbf{PCBA} \\
        
        \midrule[0.1pt]
        Linear
        & \parbox[c]{1cm}{\centering  71.24 \\ \tiny $\pm$ 0.20}
        & \parbox[c]{1cm}{\centering  82.35 \\ \tiny $\pm$ 0.14}
        & \parbox[c]{1cm}{\centering  66.28 \\ \tiny $\pm$ 0.10}
        & \parbox[c]{1cm}{\centering  72.52 \\ \tiny $\pm$ 0.08}
        & \parbox[c]{1cm}{\centering  69.54 \\ \tiny $\pm$ 0.12}
        & \parbox[c]{1cm}{\centering  82.25 \\ \tiny $\pm$ 0.23}
        & \parbox[c]{1cm}{\centering  63.71 \\ \tiny $\pm$ 0.19}
        & \parbox[c]{1cm}{\centering  56.57 \\ \tiny $\pm$ 0.24}

        \\
       
        GCN~\cite{kipf2016gcn}
        & \parbox[c]{1cm}{\centering 74.33 \\ \tiny $\pm$ 0.23}
        & \parbox[c]{1cm}{\centering 82.07 \\ \tiny $\pm$ 0.25}
        & \parbox[c]{1cm}{\centering \textcolor{blue}{69.43} \\ \tiny \textcolor{blue}{$\pm$ 0.41}}
        & \parbox[c]{1cm}{\centering \textcolor{blue}{75.26} \\ \tiny \textcolor{blue}{$\pm$ 0.22}}
        & \parbox[c]{1cm}{\centering 70.11 \\ \tiny $\pm$ 0.15}
        & \parbox[c]{1cm}{\centering 81.71 \\ \tiny $\pm$ 0.30}
        & \parbox[c]{1cm}{\centering 64.34 \\ \tiny $\pm$ 0.26}
        & \parbox[c]{1cm}{\centering 63.34 \\ \tiny $\pm$ 0.28}
        \\

        GAT~\cite{velivckovic2017gat}
        & \parbox[c]{1cm}{\centering  \textcolor{blue}{75.02} \\ \tiny \textcolor{blue}{$\pm$ 0.29}}
        & \parbox[c]{1cm}{\centering  82.46 \\ \tiny $\pm$ 0.33}
        & \parbox[c]{1cm}{\centering  68.50 \\ \tiny $\pm$ 0.21}
        & \parbox[c]{1cm}{\centering  74.92 \\ \tiny $\pm$ 0.15}
        & \parbox[c]{1cm}{\centering  71.21 \\ \tiny $\pm$ 0.25}
        & \parbox[c]{1cm}{\centering  \textcolor{blue}{82.44} \\ \tiny \textcolor{blue}{$\pm$ 0.37}}
        & \parbox[c]{1cm}{\centering  65.28 \\ \tiny $\pm$ 0.16}
        & \parbox[c]{1cm}{\centering  66.22 \\ \tiny $\pm$ 0.24}
        \\
        
        GraphSAGE~\cite{hamilton2017graphsage}
        & \parbox[c]{1cm}{\centering  74.57 \\ \tiny $\pm$ 0.34}
        & \parbox[c]{1cm}{\centering  \textcolor{blue}{82.52} \\ \tiny \textcolor{blue}{$\pm$ 0.28}}
        & \parbox[c]{1cm}{\centering  68.71 \\ \tiny $\pm$ 0.30}
        & \parbox[c]{1cm}{\centering  75.15 \\ \tiny $\pm$ 0.24}
        & \parbox[c]{1cm}{\centering  \textcolor{blue}{72.16} \\ \tiny \textcolor{blue}{$\pm$ 0.13}}
        & \parbox[c]{1cm}{\centering  81.52 \\ \tiny $\pm$ 0.24}
        & \parbox[c]{1cm}{\centering  64.56 \\ \tiny $\pm$ 0.27}
        & \parbox[c]{1cm}{\centering  66.72 \\ \tiny $\pm$ 0.33}
        \\

        GIN~\cite{xu2018gin}
        & \parbox[c]{1cm}{\centering  74.13 \\ \tiny $\pm$ 0.15}
        & \parbox[c]{1cm}{\centering  81.57 \\ \tiny $\pm$ 0.33}
        & \parbox[c]{1cm}{\centering  68.93 \\ \tiny $\pm$ 0.20}
        & \parbox[c]{1cm}{\centering  74.37 \\ \tiny $\pm$ 0.28}
        & \parbox[c]{1cm}{\centering  70.82 \\ \tiny $\pm$ 0.19}
        & \parbox[c]{1cm}{\centering  81.24 \\ \tiny $\pm$ 0.55}
        & \parbox[c]{1cm}{\centering  \textcolor{blue}{65.62} \\ \tiny \textcolor{blue}{$\pm$ 0.39}}
        & \parbox[c]{1cm}{\centering  \textcolor{blue}{67.50} \\ \tiny \textcolor{blue}{$\pm$ 0.23}}
        \\

        \midrule[0.1pt]

        FedAvg~\cite{mcmahan2017fedavg}
        & \parbox[c]{1cm}{\centering  75.15 \\ \tiny $\pm$ 0.18 }
        & \parbox[c]{1cm}{\centering  83.28 \\ \tiny $\pm$ 0.31}
        & \parbox[c]{1cm}{\centering  71.03 \\ \tiny $\pm$ 0.25}
        & \parbox[c]{1cm}{\centering  76.22 \\ \tiny $\pm$ 0.29}
        & \parbox[c]{1cm}{\centering  71.45 \\ \tiny $\pm$ 0.27}
        & \parbox[c]{1cm}{\centering  82.52 \\ \tiny $\pm$ 0.23}
        & \parbox[c]{1cm}{\centering  65.24 \\ \tiny $\pm$ 0.13}
        & \parbox[c]{1cm}{\centering  68.92 \\ \tiny $\pm$ 0.42}
        \\
       
        MOON~\cite{li2021moon}
        & \parbox[c]{1cm}{\centering  75.34 \\ \tiny $\pm$ 0.17}
        & \parbox[c]{1cm}{\centering  83.72 \\ \tiny $\pm$ 0.22}
        & \parbox[c]{1cm}{\centering  72.12 \\ \tiny $\pm$ 0.33}
        & \parbox[c]{1cm}{\centering  76.15 \\ \tiny $\pm$ 0.09}
        & \parbox[c]{1cm}{\centering  71.20\\ \tiny  $\pm$ 0.23}
        & \parbox[c]{1cm}{\centering  82.31 \\ \tiny $\pm$ 0.25}
        & \parbox[c]{1cm}{\centering  65.56 \\ \tiny $\pm$ 0.45}
        & \parbox[c]{1cm}{\centering  69.11 \\ \tiny $\pm$ 0.10}
        \\

        FedSage+~\cite{zhang2021fedsage}
        & \parbox[c]{1cm}{\centering  76.40 \\ \tiny $\pm$ 0.33}
        & \parbox[c]{1cm}{\centering  \textcolor{blue}{85.35} \\ \tiny \textcolor{blue}{$\pm$ 0.27}}
        & \parbox[c]{1cm}{\centering  73.33 \\ \tiny $\pm$ 0.48}
        & \parbox[c]{1cm}{\centering  76.22 \\ \tiny $\pm$ 0.36}
        & \parbox[c]{1cm}{\centering  \textcolor{blue}{72.95} \\ \tiny \textcolor{blue}{$\pm$ 0.29}}
        & \parbox[c]{1cm}{\centering  \textcolor{blue}{84.26} \\ \tiny \textcolor{blue}{$\pm$ 0.26}}
        & \parbox[c]{1cm}{\centering \textit{N/A}}
        & \parbox[c]{1cm}{\centering \textit{N/A}}
        \\
        
        Fed-PUB~\cite{baek2022fedpub}
        & \parbox[c]{1cm}{\centering  76.58 \\ \tiny $\pm$ 0.32}
        & \parbox[c]{1cm}{\centering  84.32 \\ \tiny $\pm$ 0.19}
        & \parbox[c]{1cm}{\centering  73.19 \\ \tiny $\pm$ 0.22}
        & \parbox[c]{1cm}{\centering  \textcolor{blue}{77.20} \\ \tiny \textcolor{blue}{$\pm$ 0.24}}
        & \parbox[c]{1cm}{\centering  72.53 \\ \tiny $\pm$ 0.41}
        & \parbox[c]{1cm}{\centering  83.48 \\ \tiny $\pm$ 0.25}
        & \parbox[c]{1cm}{\centering \textit{N/A}}
        & \parbox[c]{1cm}{\centering \textit{N/A}}
        \\

        FedGTA~\cite{li2024fedgta}
        & \parbox[c]{1cm}{\centering  76.32 \\ \tiny $\pm$ 0.44}
        & \parbox[c]{1cm}{\centering  83.91 \\ \tiny $\pm$ 0.19}
        & \parbox[c]{1cm}{\centering  73.20 \\ \tiny $\pm$ 0.22}
        & \parbox[c]{1cm}{\centering  76.86 \\ \tiny $\pm$ 0.24}
        & \parbox[c]{1cm}{\centering \textit{N/A}}
        & \parbox[c]{1cm}{\centering \textit{N/A}}
        & \parbox[c]{1cm}{\centering \textit{N/A}}
        & \parbox[c]{1cm}{\centering \textit{N/A}}
        \\

        FedTAD~\cite{zhu2024fedtad}
        & \parbox[c]{1cm}{\centering  \textcolor{blue}{76.64} \\ \tiny \textcolor{blue}{$\pm$ 0.25}}
        & \parbox[c]{1cm}{\centering  84.32 \\ \tiny $\pm$ 0.18}
        & \parbox[c]{1cm}{\centering  73.52 \\ \tiny $\pm$ 0.39}
        & \parbox[c]{1cm}{\centering  76.50 \\ \tiny $\pm$ 0.22}
        & \parbox[c]{1cm}{\centering \textit{N/A}}
        & \parbox[c]{1cm}{\centering \textit{N/A}}
        & \parbox[c]{1cm}{\centering \textit{N/A}}
        & \parbox[c]{1cm}{\centering \textit{N/A}} 
        \\

        FGSSL~\cite{huangfgl_fgssl}
        & \parbox[c]{1cm}{\centering  76.44 \\ \tiny $\pm$ 0.34}
        & \parbox[c]{1cm}{\centering  84.51 \\ \tiny $\pm$ 0.29}
        & \parbox[c]{1cm}{\centering  \textcolor{blue}{73.77} \\ \tiny \textcolor{blue}{$\pm$ 0.30}}
        & \parbox[c]{1cm}{\centering  76.62 \\ \tiny $\pm$ 0.41}
        & \parbox[c]{1cm}{\centering \textit{N/A}}
        & \parbox[c]{1cm}{\centering \textit{N/A}}
        & \parbox[c]{1cm}{\centering \textit{N/A}}
        & \parbox[c]{1cm}{\centering \textit{N/A}} 
        \\
        
        FGGP~\cite{wan2024fgl_fggp}
        & \parbox[c]{1cm}{\centering  76.55 \\ \tiny $\pm$ 0.14}
        & \parbox[c]{1cm}{\centering  84.46 \\ \tiny $\pm$ 0.25}
        & \parbox[c]{1cm}{\centering  73.25 \\ \tiny $\pm$ 0.18}
        & \parbox[c]{1cm}{\centering  76.26 \\ \tiny $\pm$ 0.11}
        & \parbox[c]{1cm}{\centering \textit{N/A}}
        & \parbox[c]{1cm}{\centering \textit{N/A}}
        & \parbox[c]{1cm}{\centering \textit{N/A}}
        & \parbox[c]{1cm}{\centering \textit{N/A}} 
        \\

        GCFL+~\cite{xie2021gcfl}
        & \parbox[c]{1cm}{\centering \textit{N/A}}
        & \parbox[c]{1cm}{\centering \textit{N/A}} 
        & \parbox[c]{1cm}{\centering \textit{N/A}}
        & \parbox[c]{1cm}{\centering \textit{N/A}}
        & \parbox[c]{1cm}{\centering \textit{N/A}}
        & \parbox[c]{1cm}{\centering \textit{N/A}}
        & \parbox[c]{1cm}{\centering  \textcolor{blue}{66.52} \\ \tiny \textcolor{blue}{$\pm$ 0.24}}
        & \parbox[c]{1cm}{\centering  71.27 \\ \tiny $\pm$ 0.16}
        \\

        FedStar~\cite{tan2023fgl_fedstar}
        & \parbox[c]{1cm}{\centering \textit{N/A}}
        & \parbox[c]{1cm}{\centering \textit{N/A}} 
        & \parbox[c]{1cm}{\centering \textit{N/A}}
        & \parbox[c]{1cm}{\centering \textit{N/A}}
        & \parbox[c]{1cm}{\centering \textit{N/A}}
        & \parbox[c]{1cm}{\centering \textit{N/A}}
        & \parbox[c]{1cm}{\centering  66.20 \\ \tiny $\pm$ 0.28}
        & \parbox[c]{1cm}{\centering  \textcolor{blue}{71.45} \\ \tiny \textcolor{blue}{$\pm$ 0.33}}
        \\

        \midrule[0.1pt]

        OFA$^*$~\cite{gfm_ofa}
        & \parbox[c]{1cm}{\centering  77.10 \\ \tiny $\pm$ 0.18}
        & \parbox[c]{1cm}{\centering  84.82 \\ \tiny $\pm$ 0.34}
        & \parbox[c]{1cm}{\centering  72.72 \\ \tiny $\pm$ 0.22}
        & \parbox[c]{1cm}{\centering  77.38 \\ \tiny $\pm$ 0.29}
        & \parbox[c]{1cm}{\centering  72.41 \\ \tiny $\pm$ 0.30}
        & \parbox[c]{1cm}{\centering  84.42 \\ \tiny $\pm$ 0.33}
        & \parbox[c]{1cm}{\centering  \textcolor{blue}{67.34} \\ \tiny \textcolor{blue}{$\pm$ 0.20}}
        & \parbox[c]{1cm}{\centering  73.24 \\ \tiny $\pm$ 0.27}
        \\

        GFT$^*$~\cite{gfm_gft}
        & \parbox[c]{1cm}{\centering  76.82 \\ \tiny $\pm$ 0.25}
        & \parbox[c]{1cm}{\centering  85.10 \\ \tiny $\pm$ 0.16}
        & \parbox[c]{1cm}{\centering  \textcolor{blue}{73.60} \\ \tiny \textcolor{blue}{$\pm$ 0.14}}
        & \parbox[c]{1cm}{\centering  \textcolor{blue}{77.52} \\ \tiny \textcolor{blue}{$\pm$ 0.22}}
        & \parbox[c]{1cm}{\centering  72.20 \\ \tiny $\pm$ 0.23}
        & \parbox[c]{1cm}{\centering  84.44 \\ \tiny $\pm$ 0.37}
        & \parbox[c]{1cm}{\centering  66.45 \\ \tiny $\pm$ 0.23}
        & \parbox[c]{1cm}{\centering  72.52 \\ \tiny $\pm$ 0.18}
        \\

        UniGraph$^*$~\cite{gfm_unigraph}
        & \parbox[c]{1cm}{\centering  77.35 \\ \tiny $\pm$ 0.41}
        & \parbox[c]{1cm}{\centering  85.52 \\ \tiny $\pm$ 0.19}
        & \parbox[c]{1cm}{\centering  73.25 \\ \tiny $\pm$ 0.25}
        & \parbox[c]{1cm}{\centering  75.90 \\ \tiny $\pm$ 0.23}
        & \parbox[c]{1cm}{\centering  72.42 \\ \tiny $\pm$ 0.18}
        & \parbox[c]{1cm}{\centering  84.66 \\ \tiny $\pm$ 0.30}
        & \parbox[c]{1cm}{\centering  67.29 \\ \tiny $\pm$ 0.33}
        & \parbox[c]{1cm}{\centering  73.16 \\ \tiny $\pm$ 0.10}
        \\

        GQT$^*$~\cite{gfm_gqt}
        & \parbox[c]{1cm}{\centering  \textcolor{blue}{77.51} \\ \tiny \textcolor{blue}{$\pm$ 0.20}}
        & \parbox[c]{1cm}{\centering  \textcolor{blue}{85.82} \\ \tiny \textcolor{blue}{$\pm$ 0.33}}
        & \parbox[c]{1cm}{\centering  72.36 \\ \tiny $\pm$ 0.15}
        & \parbox[c]{1cm}{\centering  76.18 \\ \tiny $\pm$ 0.21}
        & \parbox[c]{1cm}{\centering  \textcolor{blue}{73.15} \\ \tiny \textcolor{blue}{$\pm$ 0.28}}
        & \parbox[c]{1cm}{\centering  \textcolor{blue}{84.70} \\ \tiny \textcolor{blue}{$\pm$ 0.44}}
        & \parbox[c]{1cm}{\centering  67.26 \\ \tiny $\pm$ 0.47}
        & \parbox[c]{1cm}{\centering  73.12 \\ \tiny $\pm$ 0.23}
        \\

        GraphCLIP$^*$~\cite{gfm_graphclip}
        & \parbox[c]{1cm}{\centering  77.28 \\ \tiny $\pm$ 0.12}
        & \parbox[c]{1cm}{\centering  85.62 \\ \tiny $\pm$ 0.14}
        & \parbox[c]{1cm}{\centering  73.29 \\ \tiny $\pm$ 0.23}
        & \parbox[c]{1cm}{\centering  76.47 \\ \tiny $\pm$ 0.42}
        & \parbox[c]{1cm}{\centering  73.10 \\ \tiny $\pm$ 0.46}
        & \parbox[c]{1cm}{\centering  83.59 \\ \tiny $\pm$ 0.30}
        & \parbox[c]{1cm}{\centering  66.46 \\ \tiny $\pm$ 0.16}
        & \parbox[c]{1cm}{\centering  \textcolor{blue}{73.58} \\ \tiny \textcolor{blue}{$\pm$ 0.26}}
        \\

        \midrule[0.1pt]
        FedGFM+~\cite{zhu2025fedgfm}
        & \parbox[c]{1cm}{\centering  \textcolor{blue}{79.45} \\ \tiny \textcolor{blue}{$\pm$ 0.45}}
        & \parbox[c]{1cm}{\centering  \textcolor{blue}{87.18} \\ \tiny \textcolor{blue}{$\pm$ 0.29}}
        & \parbox[c]{1cm}{\centering  \textcolor{blue}{75.30} \\ \tiny \textcolor{blue}{$\pm$ 0.32}}
        & \parbox[c]{1cm}{\centering  \textcolor{blue}{78.26} \\ \tiny \textcolor{blue}{$\pm$ 0.20}}
        & \parbox[c]{1cm}{\centering  \textcolor{blue}{74.58} \\ \tiny \textcolor{blue}{$\pm$ 0.15}}
        & \parbox[c]{1cm}{\centering  \textcolor{blue}{85.42} \\ \tiny \textcolor{blue}{$\pm$ 0.31}}
        & \parbox[c]{1cm}{\centering  \textcolor{blue}{68.24} \\ \tiny \textcolor{blue}{$\pm$ 0.28}}
        & \parbox[c]{1cm}{\centering  \textcolor{blue}{76.50} \\ \tiny \textcolor{blue}{$\pm$ 0.16}}
        \\

        \textbf{FedBook (Ours)}
        & \parbox[c]{1cm}{\centering  \textcolor{red}{81.32} \\ \tiny  \textcolor{red}{$\pm$ 0.25}}
        & \parbox[c]{1cm}{\centering  \textcolor{red}{88.54} \\ \tiny \textcolor{red}{$\pm$ 0.14}}
        & \parbox[c]{1cm}{\centering  \textcolor{red}{76.20} \\ \tiny \textcolor{red}{$\pm$ 0.33}}
        & \parbox[c]{1cm}{\centering  \textcolor{red}{79.85} \\ \tiny \textcolor{red}{$\pm$ 0.18}}
        & \parbox[c]{1cm}{\centering  \textcolor{red}{75.82} \\ \tiny \textcolor{red}{$\pm$ 0.44}}
        & \parbox[c]{1cm}{\centering  \textcolor{red}{87.23} \\ \tiny \textcolor{red}{$\pm$ 0.26}}
        & \parbox[c]{1cm}{\centering  \textcolor{red}{69.82} \\ \tiny \textcolor{red}{$\pm$ 0.19}}
        & \parbox[c]{1cm}{\centering  \textcolor{red}{77.24} \\ \tiny \textcolor{red}{$\pm$ 0.28}} \\
        
        \bottomrule[1pt] 
    \end{tabular}
    }}
\end{table*}

\subsection{Performance Comparison (Answers for Q1)}
\label{sec: performance comparison}

To address \textbf{Q1}, we compare the performance of our proposed FedBook framework against various baselines in Table~\ref{tab: compare baseline}. FedBook consistently achieves superior performance over all baselines across diverse domains and tasks. We now provide a detailed analysis of the reasons for the subpar performance of each baseline category.

\vspace{0.1cm}
\noindent \textbf{Comparison to Isolated Supervised Models.} As observed, FedBook achieves marked improvements over isolated models across all domains and tasks. Specifically, it surpasses the best baseline by an average of 5.9\% in node classification, 4.2\% in edge classification, and 6.9\% in graph classification. This consistent advantage underscores the substantial performance gain of our method. The advancement stems primarily from the inherent benefits of the FedGFM paradigm: clients within the same domain benefit from mutual enhancement through federated optimization, thereby overcoming the limitations of isolated learning; moreover, the framework effectively captures transferable knowledge patterns across diverse domains, leading to more robust and generalized performance.

\vspace{0.1cm}
\noindent \textbf{Comparison to FL/FGL Methods.} As shown in our experiments, FedBook achieves consistent performance gains over FL/FGL baselines across multiple tasks, with average improvements of 3.2\% in node classification, 2.9\% in edge classification, and 4.5\% in graph classification. This advantage can be largely attributed to the inherent constraints of conventional FL/FGL methods, which are typically tailored to specific graph types or tasks. Such specialization limits their applicability to diverse graph tasks and further hinders their ability to capture transferable knowledge across domains in FedGFM scenarios. In contrast, FedBook can effectively integrate multi-domain and multi-task collective intelligence, thereby achieving satisfactory downstream performance across all tasks.

\noindent \textbf{Comparison to Federated Adaptations of Centralized GFM Methods.} As observed, FedBook achieves a marked improvement over federated adaptations of centralized GFM methods, surpassing the best baseline by an average of 2.9\% in node classification, 2.5\% in edge classification, and 3.0\% in graph classification. The primary limitation of these baselines lies in their straightforward aggregation process for global GFM (i.e., the server relies solely on FedAvg~\cite{mcmahan2017fedavg} for aggregation), which fails to establish the two pivotal properties essential for a robust global GFM (i.e., intra-domain collaboration and inter-domain integration). Consequently, the resulting global model remains sub-optimal, whose deficiencies become apparent during downstream task-specific fine-tuning.

\vspace{0.1cm}
    \noindent \textbf{Comparison to FedGFM Method.} FedGFM+ stands as the only existing baseline in the FedGFM paradigm. As observed, compared with the other three baseline categories, FedGFM+ delivers competitive yet consistently sub-optimal performance. In contrast, our proposed FedBook demonstrates a clear and systematic advantage, achieving performance gains of 1.4\% in node classification, 1.5\% in edge classification, and 1.1\% in graph classification. These results collectively affirm that a carefully designed aggregation process is essential to unlocking the potential of the FedGFM paradigm.

\begin{table*}[htbp]
    \setlength{\abovecaptionskip}{0.2cm} 
    \renewcommand{\arraystretch}{1.5} 
    \caption{Ablation study results for FedBook on 8 datasets across diverse domains and tasks.}
    \footnotesize 
    \label{tab: ablation study}
    \resizebox{\linewidth}{13mm}{  
    \setlength{\tabcolsep}{2.8mm}{  
    \begin{tabular}{c|c|c|c|c|c|c|c|c}
        \toprule[1pt] 
         \textbf{Method} & \textbf{Cora} & \textbf{PubMed} & \textbf{OGB-arxiv} & \textbf{WikiCS} & \textbf{FB15K237} & \textbf{WN18RR} & \textbf{HIV} & \textbf{PCBA} \\
        
        \midrule[0.1pt]

        FedBook w/o Phase 1
        & \parbox[c]{1cm}{\centering 77.75 \\ \tiny $\pm$ 0.31}
        & \parbox[c]{1cm}{\centering 86.82 \\ \tiny $\pm$ 0.38}
        & \parbox[c]{1cm}{\centering 74.24 \\ \tiny $\pm$ 0.25}
        & \parbox[c]{1cm}{\centering 77.61 \\ \tiny $\pm$ 0.19}
        & \parbox[c]{1cm}{\centering 73.26\\ \tiny $\pm$ 0.40}
        & \parbox[c]{1cm}{\centering 84.12 \\ \tiny $\pm$ 0.28}
        & \parbox[c]{1cm}{\centering 66.86 \\ \tiny $\pm$ 0.34}
        & \parbox[c]{1cm}{\centering 74.28 \\ \tiny $\pm$ 0.23}
        \\

        FedBook w/o Phase 2
        & \parbox[c]{1cm}{\centering \textcolor{blue}{78.55} \\ \tiny \textcolor{blue}{$\pm$ 0.31}}
        & \parbox[c]{1cm}{\centering \textcolor{blue}{87.30} \\ \tiny \textcolor{blue}{$\pm$ 0.44}}
        & \parbox[c]{1cm}{\centering \textcolor{blue}{75.28} \\ \tiny \textcolor{blue}{$\pm$ 0.18}}
        & \parbox[c]{1cm}{\centering \textcolor{blue}{78.50} \\ \tiny \textcolor{blue}{$\pm$ 0.22}}
        & \parbox[c]{1cm}{\centering \textcolor{blue}{74.49} \\ \tiny \textcolor{blue}{$\pm$ 0.17}}
        & \parbox[c]{1cm}{\centering \textcolor{blue}{84.96} \\ \tiny \textcolor{blue}{$\pm$ 0.35}}
        & \parbox[c]{1cm}{\centering \textcolor{blue}{67.80} \\ \tiny \textcolor{blue}{$\pm$ 0.26}}
        & \parbox[c]{1cm}{\centering \textcolor{blue}{76.33} \\ \tiny \textcolor{blue}{$\pm$ 0.28}}
        \\

        \midrule[0.1pt]
        \textbf{FedBook (Ours)}
        & \parbox[c]{1cm}{\centering  \textcolor{red}{81.32} \\ \tiny  \textcolor{red}{$\pm$ 0.25}}
        & \parbox[c]{1cm}{\centering  \textcolor{red}{88.54} \\ \tiny \textcolor{red}{$\pm$ 0.14}}
        & \parbox[c]{1cm}{\centering  \textcolor{red}{76.20} \\ \tiny \textcolor{red}{$\pm$ 0.33}}
        & \parbox[c]{1cm}{\centering  \textcolor{red}{79.85} \\ \tiny \textcolor{red}{$\pm$ 0.18}}
        & \parbox[c]{1cm}{\centering  \textcolor{red}{75.82} \\ \tiny \textcolor{red}{$\pm$ 0.44}}
        & \parbox[c]{1cm}{\centering  \textcolor{red}{87.23} \\ \tiny \textcolor{red}{$\pm$ 0.26}}
        & \parbox[c]{1cm}{\centering  \textcolor{red}{69.82} \\ \tiny \textcolor{red}{$\pm$ 0.19}}
        & \parbox[c]{1cm}{\centering  \textcolor{red}{77.24} \\ \tiny \textcolor{red}{$\pm$ 0.28}}
        \\
        \bottomrule[1pt] 
    \end{tabular}
    }}
\end{table*}

\begin{table*}[htbp]
    \setlength{\abovecaptionskip}{0.2cm} 
    \renewcommand{\arraystretch}{1.5} 
    \caption{2-shot learning results for FedBook, federated adaptations of centralized GFMs, and FedGFM+.}
    \footnotesize 
    \label{tab: few shot}
    \resizebox{\linewidth}{24mm}{  
    \setlength{\tabcolsep}{4.2mm}{  
    \begin{tabular}{c|c|c|c|c|c|c}
        \toprule[1pt] 
         \textbf{Method} & \textbf{Cora} & \textbf{PubMed} & \textbf{OGB-arxiv} & \textbf{WikiCS} & \textbf{FB15K237} & \textbf{WN18RR} \\
        
        \midrule[0.1pt]

        OFA$^*$~\cite{gfm_ofa}
        & \parbox[c]{1cm}{\centering  50.25 \\ \tiny $\pm$ 0.30}
        & \parbox[c]{1cm}{\centering  47.55 \\ \tiny $\pm$ 0.16}
        & \parbox[c]{1cm}{\centering  19.61 \\ \tiny $\pm$ 0.19}
        & \parbox[c]{1cm}{\centering  39.10 \\ \tiny $\pm$ 0.15}
        & \parbox[c]{1cm}{\centering  19.63 \\ \tiny $\pm$ 0.22}
        & \parbox[c]{1cm}{\centering  29.45 \\ \tiny $\pm$ 0.20}
        \\

        GFT$^*$~\cite{gfm_gft}
        & \parbox[c]{1cm}{\centering  \textcolor{blue}{53.22} \\ \tiny \textcolor{blue}{$\pm$ 0.24}}
        & \parbox[c]{1cm}{\centering  48.96 \\ \tiny $\pm$ 0.27}
        & \parbox[c]{1cm}{\centering  20.52 \\ \tiny $\pm$ 0.35}
        & \parbox[c]{1cm}{\centering  \textcolor{blue}{40.17} \\ \tiny \textcolor{blue}{$\pm$ 0.36}}
        & \parbox[c]{1cm}{\centering  18.74 \\ \tiny $\pm$ 0.38}
        & \parbox[c]{1cm}{\centering  28.26 \\ \tiny $\pm$ 0.31}
        \\

        UniGraph$^*$~\cite{gfm_unigraph}
        & \parbox[c]{1cm}{\centering  49.42 \\ \tiny $\pm$ 0.21}
        & \parbox[c]{1cm}{\centering  47.44 \\ \tiny $\pm$ 0.41}
        & \parbox[c]{1cm}{\centering  18.74 \\ \tiny $\pm$ 0.36}
        & \parbox[c]{1cm}{\centering  38.72 \\ \tiny $\pm$ 0.19}
        & \parbox[c]{1cm}{\centering  \textcolor{blue}{20.52} \\ \tiny \textcolor{blue}{$\pm$ 0.42}}
        & \parbox[c]{1cm}{\centering  29.70 \\ \tiny $\pm$ 0.24}
        \\

        GQT$^*$~\cite{gfm_gqt}
        & \parbox[c]{1cm}{\centering  52.46 \\ \tiny $\pm$ 0.48}
        & \parbox[c]{1cm}{\centering  48.35 \\ \tiny $\pm$ 0.15}
        & \parbox[c]{1cm}{\centering  \textcolor{blue}{21.38} \\ \tiny \textcolor{blue}{$\pm$ 0.20}}
        & \parbox[c]{1cm}{\centering  39.44 \\ \tiny $\pm$ 0.34}
        & \parbox[c]{1cm}{\centering  19.60 \\ \tiny $\pm$ 0.15}
        & \parbox[c]{1cm}{\centering  \textcolor{blue}{30.22} \\ \tiny \textcolor{blue}{$\pm$ 0.28}}
        \\

        GraphCLIP$^*$~\cite{gfm_graphclip}
        & \parbox[c]{1cm}{\centering  51.72 \\ \tiny $\pm$ 0.44}
        & \parbox[c]{1cm}{\centering  \textcolor{blue}{50.15} \\ \tiny \textcolor{blue}{$\pm$ 0.22}}
        & \parbox[c]{1cm}{\centering  20.84 \\ \tiny $\pm$ 0.17}
        & \parbox[c]{1cm}{\centering  38.59 \\ \tiny $\pm$ 0.30}
        & \parbox[c]{1cm}{\centering  20.47 \\ \tiny $\pm$ 0.19}
        & \parbox[c]{1cm}{\centering  29.54 \\ \tiny $\pm$ 0.44}
        \\

        \midrule[0.1pt]
        FedGFM+~\cite{zhu2025fedgfm}
        & \parbox[c]{1cm}{\centering  \textcolor{blue}{54.28} \\ \tiny \textcolor{blue}{$\pm$ 0.17}}
        & \parbox[c]{1cm}{\centering  \textcolor{blue}{52.14} \\ \tiny \textcolor{blue}{$\pm$ 0.46}}
        & \parbox[c]{1cm}{\centering  \textcolor{blue}{23.25} \\ \tiny \textcolor{blue}{$\pm$ 0.24}}
        & \parbox[c]{1cm}{\centering  \textcolor{blue}{42.58} \\ \tiny \textcolor{blue}{$\pm$ 0.30}}
        & \parbox[c]{1cm}{\centering  \textcolor{blue}{22.40} \\ \tiny \textcolor{blue}{$\pm$ 0.39}}
        & \parbox[c]{1cm}{\centering  \textcolor{blue}{32.42} \\ \tiny \textcolor{blue}{$\pm$ 0.24}}\\

        \textbf{FedBook (Ours)}
        & \parbox[c]{1cm}{\centering  \textcolor{red}{57.35} \\ \tiny  \textcolor{red}{$\pm$ 0.41}}
        & \parbox[c]{1cm}{\centering  \textcolor{red}{55.72} \\ \tiny \textcolor{red}{$\pm$ 0.36}}
        & \parbox[c]{1cm}{\centering  \textcolor{red}{24.41} \\ \tiny \textcolor{red}{$\pm$ 0.28}}
        & \parbox[c]{1cm}{\centering  \textcolor{red}{44.62} \\ \tiny \textcolor{red}{$\pm$ 0.35}}
        & \parbox[c]{1cm}{\centering  \textcolor{red}{25.17} \\ \tiny \textcolor{red}{$\pm$ 0.21}}
        & \parbox[c]{1cm}{\centering  \textcolor{red}{33.60} \\ \tiny \textcolor{red}{$\pm$ 0.14}}\\
        \bottomrule[1pt] 
    \end{tabular}
    }}
\end{table*}

\subsection{Ablation Study (Answer for Q2)}
\label{sec: ablation study}

To address \textbf{Q2}, we analyze FedBook's architecture by examining its two core components: Intra-domain Collaboration (Sec.~\ref{sec: Intra-Domain}), which enhances client coherence within domains, and Inter-domain Integration (Sec.~\ref{sec: Inter-Domain}), which preserves semantic diversity across different domains. Through comprehensive ablation studies, we evaluate their individual contributions using the following specific configurations: (1) \underline{\textit{FedBook w/o Phase 1}}, which replaces intra-domain collaboration with FedAvg aggregation to evaluate the importance of  knowledge sharing among semantically similar clients; (2) \underline{\textit{FedBook w/o Phase 2}}, which substitutes inter-domain integration with FedAvg to assess its impact on the global GFM's semantic diversity and adaptation capability. 

As demonstrated in Table~\ref{tab: ablation study}, the performance degradation observed in both ablated configurations confirms that each module contributes indispensably to FedBook's overall effectiveness. While the performance drop from removing either phase validates their individual necessity, the more pronounced degradation observed when ablating intra-domain collaboration suggests that establishing robust client coherence within domains is particularly critical.

\subsection{Sensitivity Analysis (Answer for Q3)}
\label{sec: sensitivity analysis}

To address \textbf{Q3}, we conduct a sensitivity analysis of the core hyperparameters in FedBook. As a FedGFM technique, FedBook's key innovation lies in its server-side two-phase aggregation mechanism. We thus focus on the hyperparameters related to these components.

\vspace{0.1cm}

\begin{figure}[t]
      \includegraphics[width=0.45\textwidth]{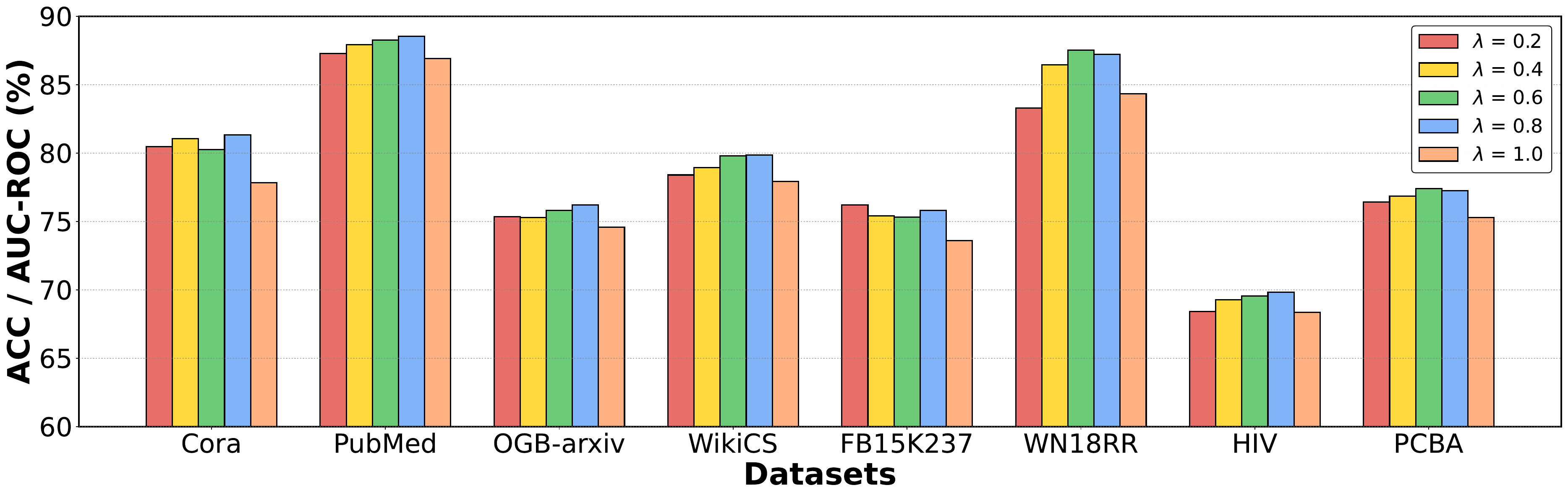}
      \captionsetup{skip=2pt, font={small}}
      \caption{Sensitivity analysis for the trade-off parameter $\lambda$.}
      \label{fig: sen_lambda}
    \end{figure}

\noindent \textbf{Sensitivity for $\mathbf{\lambda}$.} We begin by analyzing the trade-off parameter $\lambda$ in the frequency-guided semantic alignment process, which balances each token's inherent information against knowledge transferred from other tokens, i.e., Eq.~(\ref{eq: codebook_update_phase_1}). As shown in Fig. \ref{fig: sen_lambda}, varying $\lambda$ from $0.2$ to $1.0$ across eight datasets reveals that FedBook maintains stable performance despite significant parameter variations, demonstrating consistent robustness across domains and tasks. Notably, setting $\lambda = 1$, which corresponds to removing personalized codebook optimization in Phase 1, consistently degrades performance and further confirms the importance of intra-domain collaboration.

\vspace{0.1cm}

\begin{figure}[t]
      \includegraphics[width=0.45\textwidth]{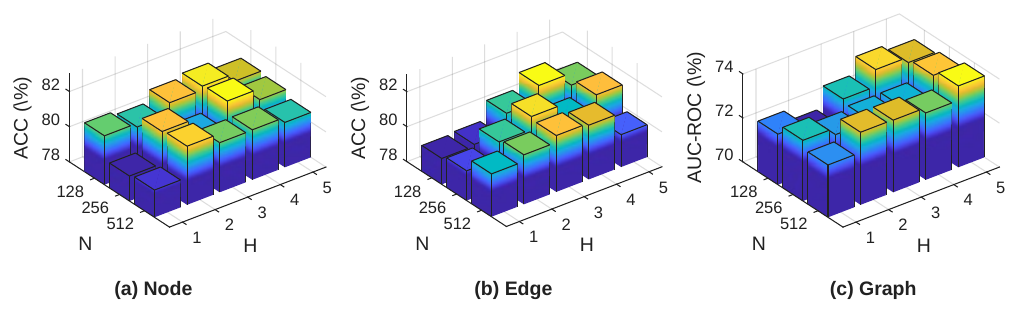}
      \captionsetup{skip=2pt, font={small}}
      \caption{Sensitivity analysis of the codebook architecture with respect to the number of heads $H$ and learnable tokens per head $N$.}
      \label{fig: sen_NH}
      \vspace{-0.1cm}
    \end{figure}

\noindent \textbf{Sensitivity for $N$ and $H$.} Moreover, we examine the influence of FedBook's backbone configuration by varying two key architectural hyperparameters: the number of codebook heads $H$ and the number of learnable tokens per head $N$. Specifically, we evaluate all combinations of $H$ ranging from 1 to 5 and $N$ in $\{128, 256, 512\}$, training the global GFM under each configuration and reporting downstream task performance across three tasks. The experimental results for node classification, edge classification and graph classification are presented in Fig.~\ref{fig: sen_NH} (a), (b) and (c), respectively. As observed, FedBook performs stably across diverse $N$ and $H$ combinations, demonstrating its holistic robustness.

\subsection{Few-Shot Analysis (Answer for Q4)}
\label{sec: few shot}
    To address \textbf{Q4}, we perform a few-shot evaluation across a range of downstream tasks. It is important to note that few-shot learning is particularly challenging on multi-label graph datasets such as HIV and PCBA. Therefore, we restrict our 2-shot evaluation to node classification and edge classification tasks, with results summarized in Table~\ref{tab: few shot}. As shown, FedBook consistently outperforms naive federated adaptations of centralized GFM training strategies across all evaluated settings. Furthermore, it also surpasses FedGFM+, a specialized federated GFM baseline, demonstrating its enhanced capability to generalize the learned global GFM to diverse downstream tasks under low-label resource scenarios.

\subsection{Efficiency Analysis (Answer for Q5)}
\label{sec: efficiency}

    \begin{figure}[t]
      \includegraphics[width=0.45\textwidth]{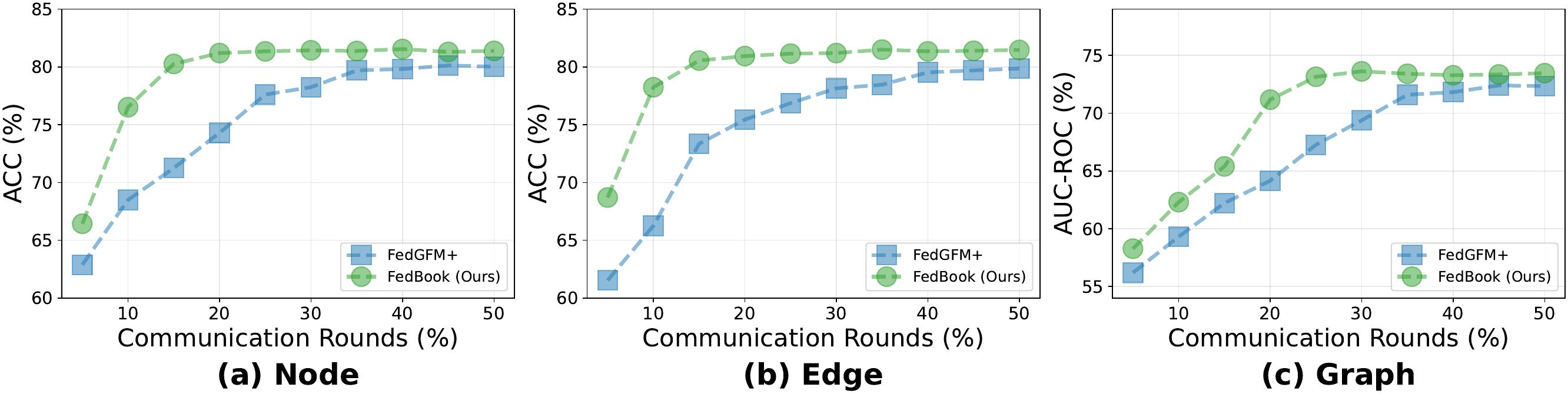}
      \captionsetup{skip=4pt, font={small}}
      \caption{Convergence curves comparing FedBook and FedGFM+, demonstrating the efficiency advantage of FedBook.}
      \label{fig: efficiency}
      \vspace{0.1cm}
    \end{figure}
    
To address \textbf{Q5}, we compare the convergence of FedBook and FedGFM+. The results in Fig.~\ref{fig: efficiency} show that across all three tasks, FedBook achieves faster convergence and superior downstream performance. Specifically, FedBook converges in 25, 30, and 30 rounds for the node, edge, and graph classification tasks, respectively, which is considerably faster than the 35, 40, and 40 rounds required by FedGFM+. This demonstrates a marked advantage for FedBook in realistic FedGFM scenarios, as it ultimately enables the pre-training of a powerful global GFM in significantly less time.

\section{Related Work}

\vspace{0.1cm}
\noindent \textbf{Federated Graph Learning (FGL).} 
 FGL adapts federated learning~\cite{Federated_Learning} to graphs by enabling decentralized, privacy-preserving GNN training across independent clients.
 From the data perspective, existing FGL studies can be grouped into three categories: (1) \underline{\textit{Graph-level FGL}}: Each client holds multiple topologically independent graphs with diverse structural and semantic characteristics. Tasks include graph-level classification and graph-level clustering. To mitigate cross-client structural heterogeneity and improve knowledge transfer, GCFL+~\cite{xie2021gcfl} and FedStar~\cite{tan2023fgl_fedstar} learn globally shared structural and semantic patterns. More studies for graph-level FGL includes FedSSP~\cite{FedSSP} and FedGCN~\cite{FedGCN_Graph_Level}.
 (2) \underline{\textit{Subgraph-level FGL}}: Each client owns a subgraph of an underlying global graph, and the loss of cross-client edges induces distribution and topology shifts. To alleviate heterogeneous data distribution, FedSage+ ~\cite{zhang2021fedsage} collaboratively train neighborhood generator to reconstruct absent neighbors. FedGL~\cite{chen2021fedgl} builds global pseudo-graph at server to complement missing local graph information. FGGP~\cite{wan2024fgl_fggp} decouples models into a feature extractor and class-wise prototypes, uploading the latter for server-side clustering to handle cross-domain shift. FedGTA~\cite{li2024fedgta} and FedTAD~\cite{zhu2024fedtad} prioritize topology-aware reliability of locally extracted node representations to improve downstream efficiency. Other works also contribute to developing the field ~\cite{zhang2024fgl_feddep,yao2024fgl_fedgcn,zhang2025Fedgm, li2024adafgl}. (3) \underline{\textit{Node-level FGL}}: Each client holds a ego subnetwork, a special case of subgraph FGL, whose local information are highly sensitive. FedEgo~\cite{zhang2023fedego} enables collaboration by transmitting mashed ego-graphs together with desensitized parameters. More insights are available in surveys \cite{fu2022fgl_survey_1, zhang2021fgl_survey_2, fu2023privacy_gml_survey} and benchmark studies \cite{he2021fedgraphnn, li2024openfgl,WangFedScope_22_fsg}.

\vspace{0.1cm}
\noindent\textbf{Graph Foundation Models(GFMs).}
GFMs~\cite{gfm_samgpt,gfm_opengraph,gfm_git,Langgfm}
typically adopt a hybrid architecture, where frozen large language models (LLMs) extract semantic features from node and edge texts, and trainable GNNs encode
topological structures of TAGs, to combine strengths of both~\cite{tang2024graphgpt,gfm_ofa,gfm_unigraph,xia2024gfm_anygraph}
so that it can learn enriched node profiles from multi-modality sources.
To further unify textual and structural knowledge, existing GFMs typically adopt self-supervised pretraining objectives such as graph contrastive learning (e.g., GraphCL~\cite{GraphCL}, GCC~\cite{GCC}) or cross-modal alignment strategies (e.g., TEA-GLM~\cite{TEA_GLM}, GraphAdapter~\cite{GraphAdapter}, GOFA~\cite{gfm_gofa}) to ensure semantic–structural consistency before fine-tuning on downstream tasks. Furthermore, A recent benchmark~\cite{GFM_Benchmark} systematically evaluates these models in the centralized setting, and provides insights on their transferability across domains such as knowledge graphs, molecular property prediction, and social network analysis.
Despite recent endeavors, research on decentralized training settings on pre-training GFMs remains limited, but it is critical considering that sensitive, text-rich graphs are distributed across numerous data holders who seek to improve GFMs' efficiency and generalizability by collaborating without sharing private raw data. 

\vspace{0.1cm}
\noindent\textbf{Federated Graph Foundation Models (FedGFMs).}
FedGFMs integrate the multi-modality strengths of GFMs with a distributive training paradigm to address decentralized and heterogeneous graph data under strict privacy constraints.
FedGFM+~\cite{zhu2025fedgfm} employs anchor-based domain initialization and a domain-sensitive prompt pool for adaptive personalization, achieving superior results.
Despite its advances, the subsequent pre-training phase still predominantly relies on simplistic parameter aggregation strategies. This overlooks the structural and semantic nuances inherent in FedGFM scenarios. Consequently, the broader challenge of systematically designing specialized and effective global codebook aggregation mechanisms remains largely unexplored.

\section{Conclusion}

This paper presents FedBook, a unified federated graph foundation codebook that addresses the challenge of achieving both intra-domain coherence and inter-domain diversity within the Federated Graph Foundation Model (FedGFM) paradigm. The FedGFM framework responds to the growing need for training graph foundation models in scenarios where data is inherently distributed across multiple clients, where data cannot be centralized due to privacy regulations, security requirements, or data sovereignty concerns. Through its carefully designed dual-phase methodology, comprising Intra-domain Collaboration for enhancing domain-specific semantics and Inter-domain Integration for maintaining heterogeneous knowledge, FedBook constructs a robust global codebook. Extensive experimental results across eight diverse benchmarks demonstrate that FedBook consistently outperforms a wide range of baseline methods, confirming its practical effectiveness in real-world FedGFM environments.

\newpage
\bibliographystyle{ACM-Reference-Format}
\balance
\bibliography{FedBook}

\appendix

\newpage
\clearpage

\section{More Experiment Details}
\label{appendix: exp setting}

\subsection{Data Processing}

Our data processing pipeline consists of two main stages: (1) \textit{Feature Extraction}: We employ Sentence-BERT~\cite{reimers2019sentence_bert} to encode textual attributes from graph datasets across diverse domains, uniformly transforming node and edge text into 768-dimensional vector representations; and (2) \textit{Decentralization Simulation}: Real-world graph data is inherently collected by multiple parties, leading to natural data decentralization. Prior FGL study~\cite{he2021fedgraphnn} categorizes this decentralization into three canonical levels: (a) node-level, where each client holds ego-networks from a global graph; (b) subgraph-level, where each client possesses a locally induced subgraph; and (c) graph-level, where each client independently collects a set of graphs from a larger collection. We note that the node-level setting is a special case of subgraph-level decentralization. Consequently, our study focuses on the latter two. To simulate these scenarios, we adopt two partitioning strategies: the Metis algorithm~\citep{karypis1998metis} for subgraph-level and random allocation for graph-level decentralization, both of which are standard in FGL literature~\cite{li2024openfgl}.

\subsection{Hyperparameters}

For \textit{Isolated Supervised Models}, models were trained locally for up to 1,000 epochs with early stopping based on validation performance. \textit{FL/FGL Baselines} were trained over 100 communication rounds with 2 local epochs per round. We used Adam optimizer with a learning rate of $1 \times 10^{-2}$, weight decay of $5 \times 10^{-4}$, and dropout rate of 0.5. For \textit{federated adaptations of centralized GFM baselines} and \textit{FedGFM+}, we adopted hyperparameters from their original papers where available; otherwise, we employed automated hyperparameter optimization via Optuna framework~\cite{akiba2019optuna}.
For our proposed \textit{FedBook}, we fixed the pre-training learning rate at $1 \times 10^{-4}$. During fine-tuning, we performed grid search over learning rates in $\{10^{-5}, 10^{-4}, 10^{-3}, 10^{-2}, 10^{-1}\}$ per dataset, with weight decay fixed at $5 \times 10^{-4}$ and batch size set to 1,024. The communication rounds for Phase 1 and Phase 2 during federated pre-tuning were searched from 1 to 50, with 2 local epochs per round maintained throughout. 

\subsection{Experimental Environment}
The experimental machine is an Intel(R) Xeon(R) Gold 6240 CPU @ 2.60GHz and NVIDIA A100 with 80GB memory and CUDA 12.4. The operating system is Ubuntu 22.04.5 with 251GB memory.

\section{Data and Code Availability}
\label{appendix: availability}

To ensure reproducibility, all datasets used in this study are publicly available. The complete source code for implementing the FedBook framework is released publicly at \url{https://anonymous.4open.science/r/FedBook-3B51}.

\end{document}